%% file: paper.tex
\documentclass[]{fairmeta}
\input{math_commands.tex}

\usepackage{hyperref}
\usepackage{url}
\usepackage{booktabs}
\usepackage{multirow}
\usepackage{array}
\usepackage{amsmath,amssymb,mathtools}
\usepackage{microtype}
\usepackage{graphicx}
\usepackage{xcolor}
\usepackage{colortbl}
\usepackage{tikz}
\usetikzlibrary{arrows.meta,positioning,calc,fit,decorations.pathreplacing}
\usepackage{algorithm}
\usepackage{float}
\usepackage{algpseudocode}
\usepackage{subcaption}
\usepackage{enumitem}
\usepackage{makecell}
\usepackage{pifont}
\usepackage{wrapfig}
\usepackage{caption}
\definecolor{rvpurple}{RGB}{95,82,170}
\definecolor{rvviolet}{RGB}{137,112,202}
\definecolor{rvteal}{RGB}{63,150,147}
\definecolor{rvorange}{RGB}{220,145,74}
\definecolor{rvgray}{RGB}{242,242,247}
\definecolor{rvline}{RGB}{105,105,120}
\definecolor{tableheader}{RGB}{237,239,243}
\definecolor{rvcorow}{RGB}{255,243,232}
\newcommand{\method}{\textsc{RVCO}}

\definecolor{DeepBlue}{RGB}{0,51,102}

\title{Residual Visual Credit Optimization: Conserved Evidence Routing for Multimodal Reinforcement Learning}

\author[1]{Lin Qiu}
\author[1]{Yao Liu}
\author[4]{Diyi Hu}
\author[2]{Hanqing Zeng}
\author[1]{Onur Gungor}
\author[3]{Chujie Chen}
\author[2]{Jiayi Liu}
\author[2]{Jianyu Wang}
\author[3]{XueLin Zheng}

\affiliation[1]{Meta Superintelligence Labs}
\affiliation[2]{Meta Recommendation System}
\affiliation[3]{Meta}
\affiliation[4]{University of Southern California}

\abstract{Reinforcement learning with verifiable rewards scales multimodal reasoning, but an outcome reward says how much a trajectory is worth, not how that value should be spread over the decisions that produced it. We introduce Residual Visual Credit Optimization (RVCO), which treats token credit as a conserved routing problem. A controlled visual intervention yields a per-token evidence response; robust within-trajectory coordinates remove incidental scale; and a budgeted entropic router distributes a fixed amount of sequence utility according to perceptual dependence. A residual support path guarantees positive credit at every valid position, and an analytic correction restores the prescribed credit mass exactly. The resulting field is selective, bounded, full-support, and invariant to response-local score shifts, and recovers hard token selection as a limiting case. Across four model families and seven reasoning benchmarks, RVCO improves accuracy over strong RLVR baselines while maintaining late-stage optimization stability, corruption robustness, and competitive training cost. Rewards, rollouts, and the group-relative advantage estimator are unchanged; only the geometry of token-level credit differs.}

\correspondence{Lin Qiu at \email{lin.qiu.stats@gmail.com}}

\begin{document}
\maketitle



\section{Introduction}
Reinforcement learning with verifiable rewards (RLVR) \citep{lightman2024verify,shao2024deepseekmath,guo2025deepseekr1} has emerged as a key mechanism for scaling reasoning in large language models. It turns automatically verified outcomes into effective policy updates without requiring dense process supervision \citep{wen2025rlvr,yu2025dapo}. The extension of RLVR to large vision-language models has shown strong potential for visual reasoning \citep{huang2025visionr1}. Mathematical diagrams, charts, documents, and structured visual questions often support objective final-answer verification \citep{meng2025mmeureka,liu2025visualrft}. Dense annotation of intermediate reasoning steps remains expensive for these tasks \citep{wang2025vlrethinker}. Modern multimodal backbones \citep{bai2023qwenvl,bai2025qwen25vl,bai2025qwen3vl} further make this regime practical at increasing context length \citep{gemini2024gemini15} and model scale \citep{zhu2025internvl3,dubey2024llama3,li2024llavaonevision}.

The central optimization problem, however, is richer than assigning a scalar score to a completed response. A multimodal trajectory is a structured computation in which visually grounded decisions, linguistic transitions, symbolic manipulations, and later conclusions remain coupled across hundreds or thousands of generated positions. Once a sequence-level utility is available, the policy still needs an allocation law that decides \emph{where} that utility enters the autoregressive computation. Uniform broadcast is simple and stable, yet it ignores heterogeneous visual dependence. Perception-aware RL therefore increasingly introduces visual interventions \citep{wang2026papo,huang2026vppo,ye2026pgpo}, confidence signals \citep{xiao2026vlcalibration}, process rewards \citep{min2026prm}, hidden-state similarity \citep{li2026pepo,wang2026vgpo}, or token-level advantage reshaping \citep{li2026prpo}. This body of work establishes token credit as an important interface between outcome supervision and grounded multimodal computation, but it leaves open what a credit operator should guarantee: existing designs fix either the active support or the score transform, and the total amount of update they inject is a consequence of that choice rather than a controlled quantity.

A visual-credit operator must reconcile three properties that are individually natural but jointly nontrivial. \emph{Selectivity} requires assigning greater credit to decisions whose likelihood depends more strongly on visual evidence. \emph{Support continuity} requires maintaining a learning signal along the surrounding reasoning trajectory, since visual evidence can influence distributed spans rather than isolated tokens \citep{sun2025visualforgetting,favero2024grounding}. \emph{Utility conservation} requires controlling the total credit allocated to a trajectory independently of response length, evidence scale, and training stage. Hard top-$\rho$ routing provides explicit selectivity and budget control but introduces discontinuous support \citep{rao2021dynamicvit,shi2023upop}; independent soft gates preserve continuous support but allow the total credit mass to vary with score calibration. Existing approaches therefore leave unresolved how to achieve selective, continuous, and budget-preserving credit within a single operator.

\begin{figure}[H]
\centering
\includegraphics[
    width=\linewidth,
    trim=0 180 0 180,
    clip
]{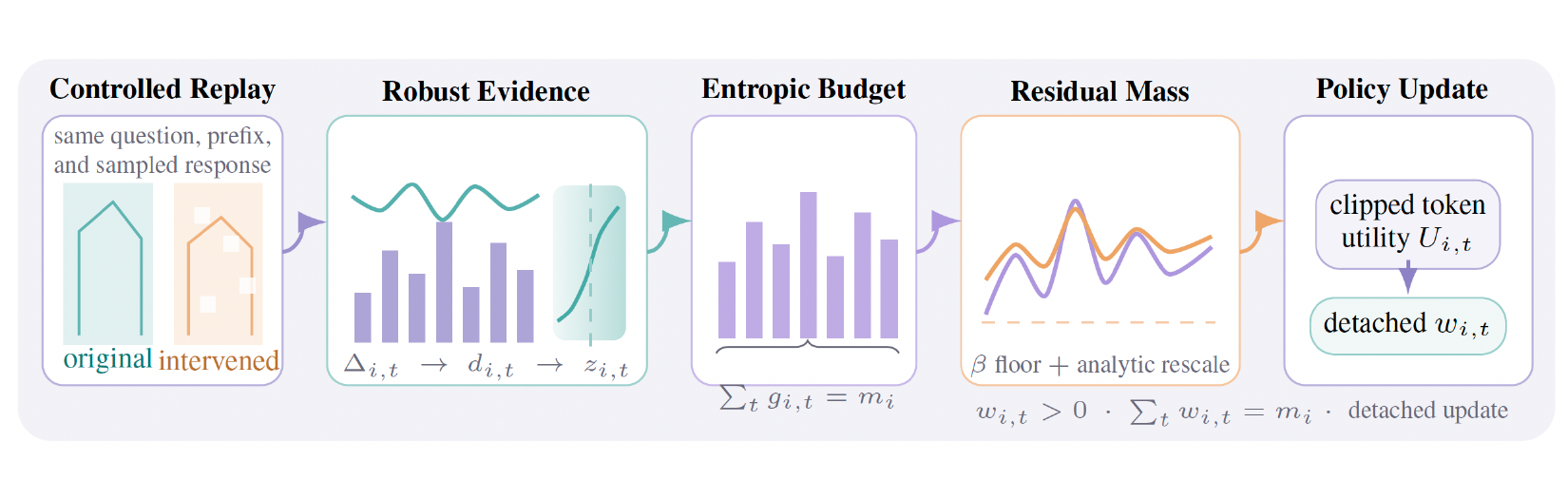}
\caption{\textbf{Conserved evidence routing in \method.}
Controlled visual replay produces a sampled-token evidence response without changing the generated trajectory.
Response-local calibration removes incidental scale, the entropic router concentrates a fixed mass $m_i$,
and residual mixing preserves positive support before analytic correction restores the exact budget.
Only the detached token weights enter policy optimization; rewards and advantages remain unchanged.}
\label{fig:intro_credit}
\end{figure}

We formulate visual credit as \emph{conserved evidence routing}. Residual Visual Credit Optimization (\method) compares each sampled token under the original image and a controlled visual intervention while holding the textual state fixed. The resulting likelihood response is transformed into a robust within-trajectory coordinate and passed through an entropy-regularized router that allocates a fixed credit budget according to visual dependence. A residual path maintains positive support for valid tokens, while a closed-form correction restores the prescribed mass exactly. The reward, rollout procedure, and group-relative advantage estimator remain unchanged; \method{} modifies only the geometry of token-level credit. Its components have distinct roles: the evidence coordinate determines where credit is allocated, the temperature $\tau$ controls concentration, the residual ratio $\beta$ controls support breadth, and the visual budget $\rho$ determines total credit mass. This yields a continuous bridge between sparse and dense credit allocation without entangling these controls in an unconstrained score transformation. We evaluate \method{} across multiple model families and scales on visual mathematics, geometry, logical and multidisciplinary reasoning, as well as corruption robustness and out-of-domain VQA.

The main contributions of this paper are summarized as follows:

\noindent\hangindent=1.25em\hangafter=1
\textbullet\hspace{0.45em}\textbf{Conserved Evidence Routing.} We formulate multimodal token credit as a constrained allocation problem that jointly enforces visual selectivity, continuous support, and a fixed response-level credit budget.

\noindent\hangindent=1.25em\hangafter=1
\textbullet\hspace{0.45em}\textbf{Residual Visual Credit Optimization.} We introduce \method{}, which combines robust intervention-based evidence, budgeted entropic routing, residual full-support credit, and exact mass correction. The resulting credit field is bounded, positive at every valid token, order-preserving, and exactly budget-preserving.

\noindent\hangindent=1.25em\hangafter=1
\textbullet\hspace{0.45em}\textbf{Evaluation across Models and Reasoning Tasks.} Across multiple VLM families and scales, \method{} consistently improves over strong RLVR and perception-aware baselines, with additional analyses demonstrating robustness to visual corruption and improved late-stage optimization stability.

\section{Related Work}
\textbf{Multimodal RLVR.} RLVR enables policy optimization from automatically verifiable outcomes without dense preference or process supervision. GRPO \citep{shao2024deepseekmath} provides a critic-free group-relative estimator, while DAPO \citep{yu2025dapo} develops a scalable online training recipe. Multimodal extensions build on this paradigm through curated reasoning data, rule-based rewards \citep{huang2025visionr1,meng2025mmeureka}, self-reflection, and visual reinforcement fine-tuning \citep{liu2025visualrft,wang2025vlrethinker}. Failure-centric resampling and correction-specific rollouts further improve the construction of informative training trajectories \citep{wang2026care,ding2026octopus}. These methods primarily improve the data, rollouts, rewards, or sequence-level optimization. In contrast, \method{} holds the sequence-level utility fixed and studies how it should be allocated across token decisions within a trajectory.

\textbf{Perceptual credit assignment.} Recent multimodal RL methods increasingly differentiate token updates according to visual dependence. PAPO \citep{wang2026papo} derives an implicit perception objective by contrasting original and corrupted visual inputs; VPPO \citep{huang2026vppo} identifies perceptually pivotal tokens and concentrates updates on a sparse subset; and PGPO \citep{ye2026pgpo} and PRPO \citep{li2026prpo} reshape token-level advantages using estimates of visual dependency. PEPO \citep{li2026pepo} combines perception priors with token entropy, VGPO \citep{wang2026vgpo} addresses temporal visual forgetting, and VL-Calibration \citep{xiao2026vlcalibration} separates visual from reasoning confidence. Process-reward and attribution methods provide complementary forms of dense supervision \citep{min2026prm,wu2026tokenreward}. Collectively, these methods establish the value of identifying \emph{where} visual evidence matters. \method{} addresses the complementary question of \emph{how} such evidence should be converted into token-level credit: it treats allocation as a constrained routing problem with explicit control over support, concentration, and total credit mass.

\textbf{Visual grounding and intervention.} Counterfactual saliency and visual contrastive decoding use controlled changes to visual input to probe or strengthen the dependence of model predictions on image evidence \citep{wang2023counterfactual,leng2024vcd,favero2024grounding}. Related benchmarks such as HallusionBench and SimpleVQA expose failures in visual grounding, illusion resistance, and multimodal factuality \citep{guan2024hallusionbench,cheng2025simplevqa}. These works motivate visual intervention as a useful diagnostic signal. \method{} uses intervention differently: rather than modifying decoding or defining an auxiliary reward, it measures the response of the \emph{sampled token likelihood} under controlled visual perturbation and uses that response only to construct a detached credit field.

\textbf{Credit geometry and optimization stability.} Token-level credit rules induce different optimization geometries even when they are constructed from similar visual-dependence signals. Hard selection provides explicit sparsity and budget control but changes support discontinuously at rank boundaries, whereas unconstrained continuous reweighting smooths token allocation while allowing the effective credit mass to vary with score calibration. Related work on dynamic token sparsification shows that changing active sets can itself become an optimization variable \citep{rao2021dynamicvit,shi2023upop}, while analyses of multimodal policy optimization emphasize the importance of late-stage stability \citep{huang2026vppo,wang2026papo}. \method{} separates three properties that are coupled in these alternatives:
the visual budget $\rho$ controls total credit mass, the gate temperature
$\tau$ controls routing concentration, and the residual ratio $\beta$
controls support. This decomposition motivates our analyses of optimization stability, response length, intervention choice, and data efficiency.


\section{Residual Visual Credit Optimization}

\subsection{Problem Setup}
For an image-question pair $x=(I,q)$, a behavior policy
$\pi_{\bar\theta}$ samples a group of $G$ responses
$y^{(i)}=(y^{(i)}_1,\ldots,y^{(i)}_{T_i})
\sim\pi_{\bar\theta}(\cdot\mid I,q)$.
A verifier returns a sequence reward $R_i$, and group-relative
normalization produces an advantage $A_i$.
Standard clipped policy optimization then broadcasts $A_i$ across all
valid positions \citep{schulman2017ppo,shao2024deepseekmath}.
\method{} instead constructs a detached token field
$w_i\in\mathbb{R}^{T_i}$ under a visual budget
$m_i=\lceil\rho T_i\rceil$.
The visual budget $\rho$ controls the total credit mass allocated to the
trajectory: it matches the mass of a top-$\rho$ sparse allocation without
requiring the support itself to be binary.

The desired operator is characterized by four response-local conditions, written compactly as
\begin{equation}
0<w_{i,t}\le 1,\qquad \sum_{t=1}^{T_i}w_{i,t}=m_i,\qquad z_{i,t}\ge z_{i,s}\Rightarrow w_{i,t}\ge w_{i,s},\qquad w_i=\mathcal{G}(z_i;\rho,\tau,\beta).
\label{eq:constraints}
\end{equation}
The first condition preserves support, the second fixes credit mass,
the third preserves evidence order, and the last keeps allocation local to the sampled trajectory. No token label, learned critic, or auxiliary reward model is introduced.

\subsection{Overview}
\begin{figure}[t]
\centering
\includegraphics[
    width=\linewidth,
    trim=0 40 0 0,
    clip
]{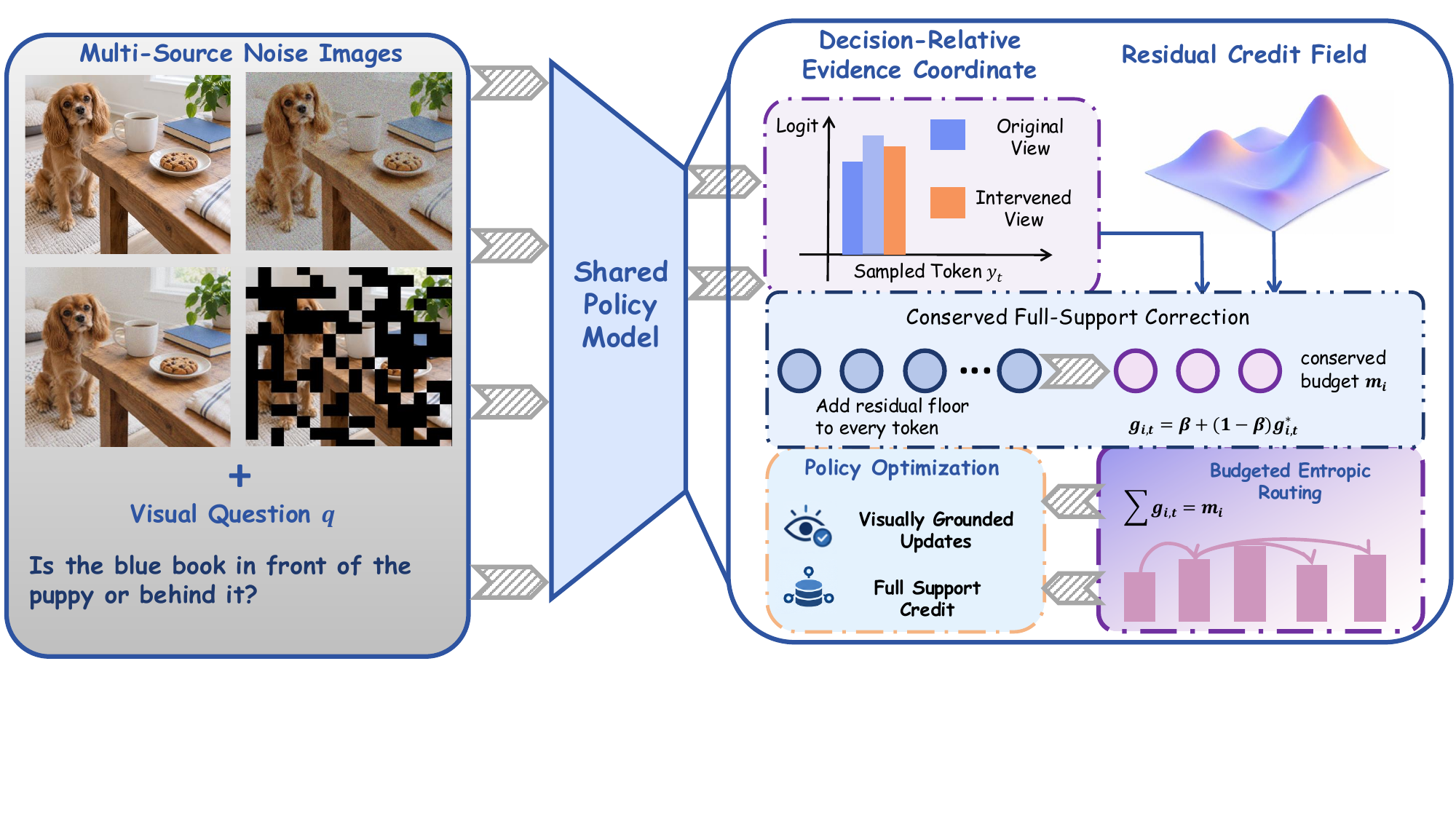}
\caption{\textbf{Main architecture of \method.} Multiple controlled visual interventions are evaluated by a shared behavior policy while the visual question and generated response state are held fixed. The resulting original/intervened token responses define a decision-relative evidence coordinate, which is transformed into a residual credit field and routed under the exact budget constraint $\sum_t g_{i,t}=m_i$. The conserved field then modulates the ordinary policy update, emphasizing visually grounded decisions while retaining full-support credit across the response.}
\label{fig:method}
\end{figure}

Figure~\ref{fig:method} shows the full computation. The sampled response is replayed under an intervened image with the question and prefix fixed, yielding one likelihood comparison per generated token. A monotone compression followed by IQR normalization produces a scale-free evidence coordinate. An entropy-regularized router then allocates exactly $m_i$ units of credit, after which residual mixing keeps every token connected to the outcome signal. Analytic mass correction removes the mass increase caused by residual mixing, and the resulting detached weights multiply the ordinary clipped token utility. The same rollout, verifier, reward, and advantage tensors can therefore be passed through an unchanged RLVR trainer.

\subsection{Visual Intervention as a Token Evidence Response}
The policy itself provides a task-adaptive probe of visual dependence. For response $i$, let $\widetilde I=\mathcal{T}(I;\xi)$ denote an image intervention that independently suppresses visual patches with probability $0.5$. The same sampled token and prefix are evaluated under both inputs,
\begin{equation}
\begin{aligned}
\ell_{i,t}
&=
\log \pi_{\bar{\theta}}
\left(
y_t^{(i)}
\mid I, q, y_{<t}^{(i)}
\right), \\
\tilde{\ell}_{i,t}
&=
\log \pi_{\bar{\theta}}
\left(
y_t^{(i)}
\mid \tilde{I}, q, y_{<t}^{(i)}
\right), \\
\Delta_{i,t}
&=
\operatorname{clip}
\left(
\tilde{\ell}_{i,t} - \ell_{i,t},
-20, 20
\right).
\end{aligned}
\label{eq:shift}
\end{equation}

Rather than retaining a signed shift, \method{} maps it to a non-negative convex response
\begin{equation}
d_{i,t}=\exp(\Delta_{i,t})-\Delta_{i,t}-1,\qquad d_{i,t}\leftarrow\operatorname{clip}(d_{i,t},0,10).
\label{eq:evidence}
\end{equation}
The function is zero when intervention leaves the sampled-token likelihood unchanged and increases for shifts in either direction. Measuring the sampled decision directly also keeps the evidence signal aligned with the action whose log-probability appears in the policy gradient, rather than summarizing changes over the entire vocabulary. Because only one scalar is retained per generated position, the construction avoids storing a full-vocabulary attribution tensor. The evidence response is detached before optimization and is not interpreted as a new reward or formal causal effect.

\subsection{Scale-Free Evidence Coordinates}
Raw likelihood responses vary with policy confidence, response composition, and training stage. Their order is often more transferable than their absolute magnitude. A monotone compression and robust response-local normalization therefore define
\begin{equation}
x_{i,t}=\log(1+d_{i,t}),\qquad
z_{i,t}=\operatorname{clip}\!\left(\frac{x_{i,t}-Q_{i,.50}}{\max(Q_{i,.75}-Q_{i,.25},\varepsilon)},-\kappa,\kappa\right),
\quad \varepsilon=10^{-6},~\kappa=8,
\label{eq:coordinate}
\end{equation}
where $Q_{i,p}$ is the empirical $p$-quantile over valid tokens. This transformation preserves within-response evidence order, is unaffected by a shared shift in $x_i$, and prevents one extreme token from setting the scale for an entire trajectory. Using the interquartile range also makes the coordinate less sensitive to a small number of unusually large intervention responses, which can occur as the policy becomes increasingly confident during training.

\subsection{Budgeted Entropic Routing}
Credit is allocated jointly rather than by independent token gates. Let $g_i\in[0,1]^{T_i}$ and $h(a)=-a\log a-(1-a)\log(1-a)$. The routed field solves
\begin{equation}
 g_i^*=\arg\max_{0\le g_i\le 1,\;\mathbf{1}^\top g_i=m_i}\Big[g_i^\top z_i+\tau\sum_{t=1}^{T_i}h(g_{i,t})\Big],\qquad
 g^*_{i,t}=\sigma\!\left(\frac{z_{i,t}-\lambda_i}{\tau}\right),\quad \sum_t g^*_{i,t}=m_i.
\label{eq:gate}
\end{equation}

The scalar dual variable $\lambda_i$ is obtained by 32 bisection steps.
For $0 < m_i < T_i$, the mass equation is continuous and strictly
decreasing in $\lambda_i$, yielding a unique finite solution.
In the boundary case $m_i = T_i$, we directly set
$g_{i,t}^{*}=1$ for all valid positions. For $0<m_i<T_i$, unlike independent gating, the equality constraint makes positions compete for the same response-level budget: increasing the allocation of one region necessarily changes how the remaining mass is distributed elsewhere. This coupling is mediated only by the scalar $\lambda_i$, keeping the routing computation simple despite the joint constraint. The gate temperature $\tau$ controls routing concentration but not total
credit mass: decreasing $\tau$ approaches a rank selector, while
increasing it spreads the same mass over a broader set of positions.

\subsection{Residual Support and Exact Conservation}
Finite-precision kernels and downstream thresholds can erase very small gate values. Positive support is therefore made explicit through residual mixing, followed by a closed-form correction:
\begin{equation}
\bar g_{i,t}=\beta+(1-\beta)g^*_{i,t},\qquad
w_{i,t}=\frac{m_i\bar g_{i,t}}{\sum_s\bar g_{i,s}}
=\frac{m_i\,[\beta+(1-\beta)g^*_{i,t}]}{\beta T_i+(1-\beta)m_i},\qquad 0<\beta<1.
\label{eq:residual}
\end{equation}
This step separates guaranteed support from selectivity.
The residual ratio $\beta$ controls the strength of the full-support
path. A small $\beta$ preserves sparse-like contrast, whereas larger
values move the field toward uniform credit without altering the
prescribed mass. Importantly, the residual path is not another smoothing
operator on $z_i$: it acts after routing and therefore ensures positive
support while preserving the evidence ordering produced by the entropic
gate.

\textbf{Proposition 1 (conserved full support).} For $0<\beta<1$ and $0<m_i\le T_i$, Equation~\ref{eq:residual} satisfies $0<w_{i,t}\le1$ and $\sum_t w_{i,t}=m_i$. It is monotone in $g^*_{i,t}$, invariant to additive shifts of $z_i$, and converges to the top-$m_i$ selector as $\tau\to0$ and $\beta\to0$ when the coordinates are distinct.

\emph{Proof sketch.} Positivity follows from $\beta>0$, while the denominator is exactly $\sum_t\bar g_{i,t}=\beta T_i+(1-\beta)m_i$. Since $m_i\le T_i$, the multiplicative correction is at most one and hence $w_{i,t}\le1$. Monotonicity is inherited from the sigmoid gate. Adding a constant to every $z_{i,t}$ shifts the dual variable by the same constant and leaves $g_i^*$ unchanged. As $\tau\to0$, the entropy term vanishes and the capped linear program selects the $m_i$ largest coordinates; $\beta\to0$ removes the residual floor. A complete derivation appears in Appendix~\ref{app:proof}.

\subsection{Full-Support Policy Optimization}
For policy $\pi_\theta$, define the importance ratio and asymmetric clipped token utility on one line,
\begin{equation}
r_{i,t}(\theta)=\frac{\pi_\theta(y^{(i)}_t\mid I,q,y^{(i)}_{<t})}{\pi_{\bar\theta}(y^{(i)}_t\mid I,q,y^{(i)}_{<t})},\qquad
U_{i,t}(\theta)=\min\!\big(r_{i,t}A_i,\operatorname{clip}(r_{i,t},1-\epsilon_l,1+\epsilon_h)A_i\big).
\label{eq:utility}
\end{equation}
The final objective is
\begin{equation}
\mathcal{L}_{\mathrm{RVCO}}(\theta)=-\frac{1}{G}\sum_{i=1}^{G}\frac{1}{T_i}\sum_{t=1}^{T_i}\operatorname{stopgrad}(w_{i,t})\,U_{i,t}(\theta).
\label{eq:objective}
\end{equation}
Because the weights sum to $m_i$, RVCO preserves the prescribed
token-credit budget while keeping every valid token connected to
sequence utility. The stop-gradient operation is important here:
intervention-based evidence controls the relative weighting of each
token utility but does not create an additional differentiation path
through the credit-construction procedure. The additional operations are response-local quantiles, elementwise transforms, and scalar bisection. Standard attention architectures and AdamW optimization remain unchanged \citep{vaswani2017attention,loshchilov2019adamw}.

\textbf{Overall Process of \method{}.}
\method{} converts intervention-based visual responses into a conserved token-level credit field through three steps: robust evidence calibration, budgeted entropic routing, and residual full-support correction. The resulting weights remain response-local and detached from policy optimization, leaving the verifier, group-relative advantage estimator, and PPO training pipeline unchanged. The complete optimization procedure is summarized in Algorithm~\ref{alg:rvco}.

\begin{algorithm}[H]
\caption{Residual Visual Credit Optimization}
\label{alg:rvco}
\small
\begin{algorithmic}[1]
\Require: behavior policy $\pi_{\bar\theta}$, policy $\pi_\theta$,
pair $(I,q)$, visual budget $\rho$, gate temperature $\tau$,
residual ratio $\beta$
\State Sample a rollout group $\{y^{(i)}\}_{i=1}^G$, evaluate rewards, and compute group-relative advantages $\{A_i\}$
\State Construct an intervened image $\widetilde I=\mathcal{T}(I;\xi)$ while keeping sampled responses and prefixes fixed
\For{each response $i$}
    \State Compute $\Delta_{i,t}$, $d_{i,t}$, and $z_{i,t}$ for valid tokens using Eqs.~\ref{eq:shift}--\ref{eq:coordinate}
    \State Set $m_i=\lceil\rho T_i\rceil$ and solve $\sum_t \sigma((z_{i,t}-\lambda_i)/\tau)=m_i$
    \State Form $g^*_{i,t}$ and conserved full-support weights $w_{i,t}$ using Eq.~\ref{eq:residual}
\EndFor
\State Update $\theta$ with Eq.~\ref{eq:objective}; no gradient flows through $w_{i,t}$
\end{algorithmic}
\end{algorithm}

\section{Experiments}
\subsection {Experimental settings}

\textbf{Models and training data.} Our primary policy is Qwen3-VL-8B-Instruct \citep{bai2025qwen3vl}.
Cross-family experiments use Qwen2.5-VL-7B \citep{bai2025qwen25vl},
InternVL3-8B \citep{zhu2025internvl3}, and Llama-3.2-Vision-11B
\citep{dubey2024llama3}; scale analysis additionally includes
Qwen3-VL-4B and Qwen3-VL-32B.
All models are trained on ViRL39K, a 38,870-example verifiable
multimodal reasoning corpus \citep{wang2025vlrethinker}.
We evaluate on seven reasoning benchmarks: MathVerse
\citep{zhang2024mathverse}, DynaMath \citep{zou2025dynamath},
MMK12 \citep{meng2025mmeureka}, Geo3K \citep{lu2021intergps},
VisNumBench \citep{weng2025visnumbench}, LogicVista
\citep{xiao2024logicvista}, and MMMU-Pro \citep{yue2025mmmupro}.

\textbf{Baselines.}
We compare against general RLVR methods GRPO \citep{shao2024deepseekmath}
and DAPO \citep{yu2025dapo}; multimodal RL methods Vision-R1
\citep{huang2025visionr1}, MM-Eureka \citep{meng2025mmeureka},
Visual-RFT \citep{liu2025visualrft}, and VL-Rethinker
\citep{wang2025vlrethinker}; and perception-aware objectives PAPO
\citep{wang2026papo}, VPPO \citep{huang2026vppo}, PGPO
\citep{ye2026pgpo}, PRPO \citep{li2026prpo}, VL-Calibration
\citep{xiao2026vlcalibration}, and VGPO \citep{wang2026vgpo}.
Optimization-side baselines use the same Qwen3-VL trainer, rollout
budget, verifier, and decoding configuration.
Cross-model experiments retain each backbone's native visual encoder
and tokenizer.


\textbf{Optimization and metrics.}
Unless stated otherwise, we train for two epochs with AdamW at a learning
rate of $10^{-6}$, using rollout and global batch sizes of 384 and 128,
eight responses per prompt, temperature 1.0, top-$p$ 0.99, and maximum
prompt/response lengths of 4096/8192.
We use asymmetric clipping $(\epsilon_l,\epsilon_h)=(0.20,0.28)$ and
default \method{} parameters $(\rho,\tau,\beta)=(0.4,0.12,0.05)$,
corresponding to the visual budget, gate temperature, and residual ratio.
For evaluation, we report accuracy@8 (\%), the percentage of examples
with at least one exact-match correct answer among eight sampled responses
under the official extraction rules.
We additionally report credit stability, corruption robustness, training
cost, and out-of-domain VQA metrics.
Appendix~\ref{app:setup} provides the complete protocol.


\subsection{Comparison with State-of-the-Art Baselines}

\textbf{Qwen-Family Comparison.}
We first evaluate RVCO on Qwen3-VL-4B, 8B, and 32B using the same
seven benchmarks and training protocol within each backbone.
As shown in Table~1, RVCO achieves the highest average accuracy at
all three scales, improving over VPPO by 2.1, 1.1, and 1.1 points
on the 4B, 8B, and 32B models, respectively.
The gains are distributed across benchmarks rather than driven by a
single task. At 32B, VPPO is higher on DynaMath, while RVCO achieves
the highest average and leads on the other six benchmarks.
Additional Qwen3-VL-8B baselines are reported in Appendix Table~\ref{tab:main_full}.

\begin{table}[H]
\caption{Benchmark-wise comparison across the Qwen3-VL family. All values are accuracy@8 (\%); Avg. is the macro-average over the seven benchmarks with reported results. The same rollout budget, verifier, and decoding protocol are used for all methods within each backbone. Bold and underlined values denote the best and second-best results within each backbone.}
\label{tab:qwen_main}
\centering
\scriptsize
\setlength{\tabcolsep}{1.55pt}
\begin{tabular}{llcccccccc}
\toprule
\rowcolor{tableheader}
Backbone & Method & MathV. & Dyna & MMK12 & Geo3K & VisNum & Logic & MMMU-P & Avg.\\
\midrule

\multirow{4}{*}{Qwen3-VL-4B}
& GRPO
& 66.5
& 65.8
& 72.3
& 65.0
& 23.1
& 57.8
& 39.2
& 55.7\\
& Visual-RFT
& 80.7
& \underline{75.6}
& \underline{76.3}
& 67.6
& \underline{25.4}
& 59.1
& 49.3
& \underline{62.0}\\
& VPPO
& \underline{80.9}
& 75.3
& 76.2
& \underline{67.7}
& 24.8
& \underline{59.2}
& \underline{49.5}
& 61.9\\
\rowcolor{rvcorow}
& \method{}
& \textbf{82.3}
& \textbf{78.2}
& \textbf{78.0}
& \textbf{69.7}
& \textbf{27.2}
& \textbf{61.5}
& \textbf{51.1}
& \textbf{64.0}\\
\midrule

\multirow{4}{*}{Qwen3-VL-8B}
& GRPO
& 76.5
& 75.0
& 71.2
& 64.4
& 28.7
& 56.0
& 42.8
& 59.2\\
& Visual-RFT
& 81.0
& 78.1
& 77.2
& 69.1
& \underline{29.6}
& \underline{63.0}
& 49.3
& 63.9\\
& VPPO
& \underline{83.8}
& \underline{80.0}
& \underline{80.1}
& \underline{71.9}
& 29.1
& 62.3
& \underline{52.4}
& \underline{65.7}\\
\rowcolor{rvcorow}
& \method{}
& \textbf{84.4}
& \textbf{80.4}
& \textbf{80.6}
& \textbf{72.5}
& \textbf{30.5}
& \textbf{64.1}
& \textbf{55.4}
& \textbf{66.8}\\
\midrule

\multirow{4}{*}{Qwen3-VL-32B}
& GRPO
& 79.2
& 78.4
& 82.6
& 78.9
& 31.7
& 67.8
& 42.7
& 65.9\\
& Visual-RFT
& 83.0
& 84.0
& 84.8
& 79.2
& 32.3
& 67.4
& \underline{60.2}
& 70.1\\
& VPPO
& \underline{85.0}
& \textbf{86.2}
& \underline{87.0}
& \underline{81.4}
& \underline{32.8}
& \underline{69.8}
& 59.7
& \underline{71.7}\\
\rowcolor{rvcorow}
& \method{}
& \textbf{86.9}
& \underline{85.9}
& \textbf{88.8}
& \textbf{82.1}
& \textbf{33.4}
& \textbf{70.2}
& \textbf{62.3}
& \textbf{72.8}\\
\bottomrule
\end{tabular}
\end{table}

\textbf{Cross-Family Generalization.}
To test whether the gains are specific to Qwen3-VL, we apply \method{}
to Qwen2.5-VL-7B, InternVL3-8B, and Llama-3.2-Vision-11B, retaining
each backbone's native visual encoder and tokenizer.
Table~\ref{tab:crossmodel} shows consistent gains over VPPO, improving
the macro-average by 1.1, 1.1, and 1.0 points, respectively.
The largest improvements occur on MMMU-Pro, with gains of 3.0, 3.3,
and 3.4 points.
These results show that the gains extend across heterogeneous
vision-language backbones rather than being specific to Qwen3-VL.

\subsection{Ablation Study}

\textbf{Ablation Variants.}
We ablate each component of the \method{} credit field on Qwen3-VL-8B,
holding the training data, reward, rollout budget, optimizer, and decoding
protocol fixed.
\emph{w/o Robust Coordinate} bypasses Eq.~\ref{eq:coordinate} and uses
the uncalibrated evidence response directly.
\emph{w/o Budgeted Routing} replaces the constrained gate in
Eq.~\ref{eq:gate} with independent sigmoid gates followed by mass matching.
\emph{w/o Residual Support} sets $\beta=0$ in Eq.~\ref{eq:residual},
removing the explicit full-support path.
\emph{w/o Exact Conservation} replaces the analytic correction in
Eq.~\ref{eq:residual} with unit-mean normalization, no longer preserving
the prescribed credit budget $m_i$.
The evidence response in Eq.~\ref{eq:evidence} and policy objective in
Eq.~\ref{eq:objective} remain unchanged across all variants.


\textbf{Ablation Results.}
Table~\ref{tab:ablation} shows that removing any individual component
degrades overall performance, indicating that the four components provide
complementary benefits.
Removing the robust coordinate weakens performance across most benchmarks,
while replacing budgeted routing further reduces the effectiveness of
credit allocation.
Removing residual support produces the largest degradation, highlighting
the importance of maintaining full-support credit along the reasoning
trajectory.
Replacing exact conservation also consistently reduces performance,
supporting the importance of controlling the total credit mass during
optimization.
\begin{table}[H]
\caption{Benchmark-wise comparison across vision-language model families.
All values are accuracy@8 (\%); Avg. is the macro-average over the seven benchmarks.
Each backbone retains its native visual encoder and tokenizer.
Bold and underlined values denote the best and second-best results within each backbone.}
\label{tab:crossmodel}
\centering
\scriptsize
\setlength{\tabcolsep}{1.55pt}
\begin{tabular}{llcccccccc}
\toprule
\rowcolor{tableheader}
Backbone & Method & MathV. & Dyna & MMK12 & Geo3K & VisNum & Logic & MMMU-P & Avg.\\
\midrule

\multirow{4}{*}{Qwen2.5-VL-7B}
& GRPO
& 75.0
& 72.9
& 69.5
& 62.0
& 26.4
& 54.0
& 40.7
& 57.2\\

& Visual-RFT
& 79.5
& \underline{78.2}
& 75.5
& 66.7
& 28.1
& 58.0
& 46.8
& 61.8\\

& VPPO
& \underline{82.4}
& 78.0
& \underline{78.6}
& \underline{69.6}
& \underline{28.6}
& \underline{60.4}
& \underline{50.4}
& \underline{64.0}\\

\rowcolor{rvcorow}
& \method{}
& \textbf{83.1}
& \textbf{78.5}
& \textbf{79.1}
& \textbf{70.3}
& \textbf{29.0}
& \textbf{62.3}
& \textbf{53.4}
& \textbf{65.1}\\

\midrule

\multirow{4}{*}{InternVL3-8B}
& GRPO
& 73.1
& 72.3
& 68.0
& 61.5
& 25.9
& 53.1
& 39.6
& 56.2\\

& Visual-RFT
& 77.4
& 75.2
& 73.8
& 66.0
& \underline{27.3}
& 56.9
& 45.5
& 60.3\\

& VPPO
& \underline{79.9}
& \underline{76.8}
& \underline{76.4}
& \underline{68.5}
& 27.0
& \underline{58.9}
& \underline{48.1}
& \underline{62.2}\\

\rowcolor{rvcorow}
& \method{}
& \textbf{80.3}
& \textbf{77.0}
& \textbf{76.7}
& \textbf{68.9}
& \textbf{28.2}
& \textbf{60.5}
& \textbf{51.4}
& \textbf{63.3}\\

\midrule

\multirow{4}{*}{Llama-3.2-Vision-11B}
& GRPO
& 70.5
& 69.9
& 66.5
& 59.6
& 23.8
& 51.2
& 37.5
& 54.1\\

& Visual-RFT
& 73.6
& 71.6
& 71.1
& 62.9
& 25.1
& \underline{56.1}
& 42.2
& 57.5\\

& VPPO
& \underline{76.1}
& \underline{73.2}
& \underline{73.6}
& \textbf{66.4}
& \underline{25.6}
& 55.8
& \underline{44.9}
& \underline{59.4}\\

\rowcolor{rvcorow}
& \method{}
& \textbf{76.7}
& \textbf{73.6}
& \textbf{74.2}
& \underline{66.0}
& \textbf{26.2}
& \textbf{57.6}
& \textbf{48.3}
& \textbf{60.4}\\

\bottomrule
\end{tabular}
\end{table}
\subsection{Hyperparameter Sensitivity}

To assess whether RVCO depends on a narrowly tuned routing configuration,
we independently vary the residual ratio $\beta \in \{0.01,0.03,0.05,0.07,0.09\}$,
visual budget $\rho \in \{0.2,0.3,0.4,0.5,0.6\}$, and gate temperature
$\tau \in \{0.04,0.08,0.12,0.16,0.20\}$ on Qwen3-VL-8B, holding the other
parameters at their defaults.
We report training accuracy and MathVerse validation accuracy.

\begin{table}[H]
\caption{Ablation study on Qwen3-VL-8B. Each variant changes one component while preserving the same reward, rollout budget, optimizer, and decoding protocol.}
\label{tab:ablation}
\centering
\scriptsize
\setlength{\tabcolsep}{2.25pt}
\begin{tabular}{lcccccccc}
\toprule
\rowcolor{tableheader}
Variant & MathV. & Dyna & MMK12 & Geo3K & VisNum & Logic & MMMU-P & Avg.\\
\midrule

w/o Robust Coordinate
& 83.2
& 78.7
& 79.3
& 71.3
& 29.7
& 62.9
& 53.2
& 65.5\\

w/o Budgeted Routing
& 81.7
& 77.0
& 78.5
& 71.6
& 28.9
& 61.5
& 52.1
& 64.5\\

w/o Residual Support
& 80.9
& 76.4
& 77.2
& 69.0
& 27.8
& 60.6
& 49.3
& 63.0\\

w/o Exact Conservation
& 81.9
& 77.6
& 78.1
& 70.5
& 28.6
& 61.2
& 51.7
& 64.2\\

\midrule
\rowcolor{rvcorow}
\method{}
& \textbf{84.4}
& \textbf{80.4}
& \textbf{80.6}
& \textbf{72.5}
& \textbf{30.5}
& \textbf{64.1}
& \textbf{55.4}
& \textbf{66.8}\\

\bottomrule
\end{tabular}
\end{table}


\begin{figure}[H]
\centering
\includegraphics[width=.95\linewidth]{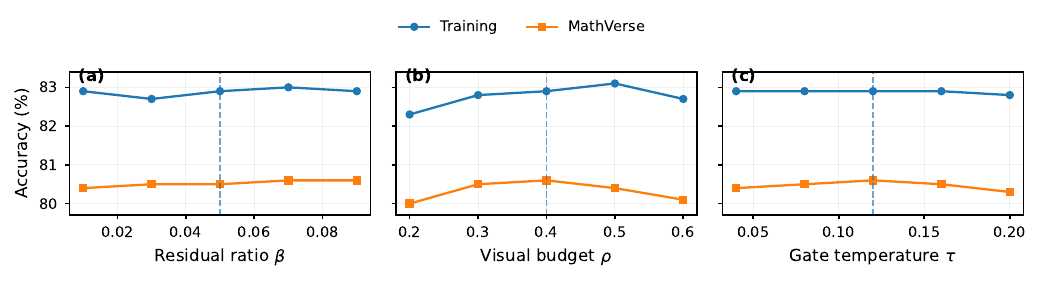}
\caption{Hyperparameter sensitivity on Qwen3-VL-8B. We vary the residual ratio $\beta$, visual budget $\rho$, and gate temperature $\tau$ while holding the other two parameters fixed. Dashed lines mark the default configuration.}
\label{fig:hypermain}
\end{figure}
Figure~\ref{fig:hypermain} shows that \method{} is stable over a
broad neighborhood of the default configuration.
MathVerse accuracy varies by only 0.2 points across the $\beta$ sweep,
while changing $\tau$ from $0.08$ to $0.20$ changes accuracy by only
0.3 points. Performance is more sensitive to the visual budget $\rho$, with $\rho=0.4$ achieving the highest validation accuracy.
These trends are consistent with the distinct roles of the three
parameters: $\rho$ controls total credit mass, $\tau$ controls routing
concentration, and $\beta$ controls residual support.


\subsection{Additional Analysis}
\textbf{Visual Robustness and Training Cost.}
We evaluate Qwen3-VL-8B under five image corruptions and report
end-to-end training cost under the same rollout budget.
For robustness, we compare GRPO, VPPO, and \method{}; for efficiency,
we report wall-clock time and GPU-hours for representative RLVR methods.
\begin{table}[H]
\caption{Visual corruption robustness and end-to-end training cost on Qwen3-VL-8B. Robustness reports exact-match accuracy (\%).}
\label{tab:robusteff}
\centering
\scriptsize
\renewcommand{\arraystretch}{1.12}

\begin{minipage}[t]{0.58\linewidth}
\centering
\textbf{Robustness}\\[1pt]

\setlength{\tabcolsep}{4.5pt}
\begin{tabular}{l@{\hspace{7pt}}ccc}
\toprule
\rowcolor{tableheader}
Corruption & GRPO & VPPO & \method{}\\
\midrule
Patch dropout (15\%) & 58.9 & 64.2 & \textbf{66.8}\\
Gaussian blur        & 57.4 & 61.9 & \textbf{64.1}\\
Low contrast         & 55.8 & 60.5 & \textbf{62.9}\\
Unseen crop          & 52.6 & 58.1 & \textbf{60.7}\\
Gaussian noise       & 54.9 & 59.8 & \textbf{62.5}\\
\bottomrule
\end{tabular}
\end{minipage}
\hfill
\begin{minipage}[t]{0.38\linewidth}
\centering
\textbf{Training Efficiency}\\[1pt]

\setlength{\tabcolsep}{5.5pt}
\begin{tabular}{l@{\hspace{9pt}}cc}
\toprule
\rowcolor{tableheader}
Method & Time & GPU-h\\
\midrule
GRPO      & 46h   & 368\\
DAPO      & 44h   & 352\\
VPPO      & 47h   & 376\\
PGPO      & 47.5h & 380\\
\rowcolor{rvcorow} \method{} & 48h   & 384\\
\bottomrule
\end{tabular}
\end{minipage}

\end{table}

Table~\ref{tab:robusteff} shows that \method{} outperforms VPPO
by 2.2--2.7 points across all five corruptions, including gains of
2.6 points on unseen crops and 2.7 points under Gaussian noise.
\method{} requires 48 hours (384 GPU-hours), compared with 47 hours
(376 GPU-hours) for VPPO.
Thus, the robustness gains are obtained with modest additional
training cost.


\textbf{End-of-Training Optimization Stability.}
We analyze the final 20\% of Qwen3-VL-8B training updates to compare
the stability of different credit geometries.
We evaluate Hard Top-$\rho$, PGPO, VPPO, entropic routing without
residual support ($\beta=0$), and the complete \method{}.
Support flicker measures changes in active credit support, while credit
continuity measures similarity between neighboring allocation fields;
we additionally report policy-entropy variance, gradient-norm
coefficient of variation, and held-out reward.



\begin{table}[H]
\caption{Optimization stability diagnostics over the final 20\% of Qwen3-VL-8B updates. Lower support flicker, entropy variance, and gradient CV are better.}
\label{tab:stability}
\centering
\small
\setlength{\tabcolsep}{4pt}
\begin{tabular}{lccccc}
\toprule
\rowcolor{tableheader}
Method & Flicker $\downarrow$ & Continuity $\uparrow$ & Ent. Var. $\downarrow$ & Grad. CV $\downarrow$ & Reward $\uparrow$\\
\midrule
Hard Top-$\rho$ & 0.219 & 0.184 & 0.047 & 0.384 & 0.751\\
PGPO & 0.143 & 0.231 & 0.031 & 0.319 & 0.766\\
VPPO & 0.126 & 0.246 & 0.024 & 0.298 & 0.758\\
Entropic routing ($\beta=0$) & 0.073 & 0.267 & 0.018 & 0.251 & 0.775\\
\rowcolor{rvcorow} \method{} & \textbf{0.041} & \textbf{0.281} & \textbf{0.012} & \textbf{0.214} & \textbf{0.784}\\
\bottomrule
\end{tabular}
\end{table}

\textbf{Stability Analysis.}
Table~\ref{tab:stability} shows a clear progression from discrete
selection to smooth routing and finally to full-support credit.
Compared with Hard Top-$\rho$, entropic routing ($\beta=0$) reduces
support flicker from 0.219 to 0.073 and increases credit continuity
from 0.184 to 0.267, indicating that smooth routing substantially
stabilizes the allocation field.
Adding residual support further reduces flicker to 0.041 and increases
continuity to 0.281.
The complete \method{} also achieves the lowest entropy variance
(0.012) and gradient CV (0.214), together with the highest held-out
reward (0.784).
These results indicate that residual support provides additional
stability beyond smoothing the routing boundary alone.



\section{Conclusion}

We introduced Residual Visual Credit Optimization (\method{}), which
formulates multimodal token credit as a conserved evidence-routing
problem. Given a fixed sequence-level utility, \method{} uses controlled visual
interventions to estimate decision-relative visual evidence and allocates
credit through budgeted entropic routing with residual full-support
correction and exact mass conservation.
The reward, rollout procedure, and group-relative advantage estimator
remain unchanged; only the token-level credit geometry is modified.
Across multiple model families and scales, \method{} improves multimodal
reasoning accuracy while maintaining robustness, optimization stability,
and modest training overhead.
These results highlight a complementary dimension of multimodal RLVR:
beyond determining how much utility a trajectory receives, effective
optimization also depends on how that utility is distributed across the
decisions that produced it.



\clearpage
\bibliographystyle{assets/plainnat}
\bibliography{paper}

\clearpage
\appendix

\input{theory}

\section{Detailed Experimental Setup}
\label{app:setup}
\subsection{Training protocol}
Each prompt produces eight candidate responses during online reinforcement learning. A rule-based verifier assigns binary accuracy rewards and response groups without reward variation are removed by online filtering. The vision tower remains trainable. Policy updates use lower and upper clipping ratios of 0.20 and 0.28, weight decay $10^{-2}$, zero warmup, maximum gradient norm 1.0, and global batch size 128. Sampling uses temperature 1.0 and top-$p$ 0.99. Prompt and response limits are 4096 and 8192 tokens. Training uses BF16, activation checkpointing, fully sharded data parallelism, and padding-free packing. The default visual intervention independently suppresses image patches with probability 0.5. Evidence responses are clipped before logarithmic compression; the robust scale is lower-bounded by $10^{-6}$ and coordinates by $[-8,8]$. All gate roots use 32 bisection iterations.

\begin{table}[h]
\caption{Additional reinforcement-learning hyperparameters.}
\label{tab:hyperparams}
\centering
\small
\begin{tabular}{ll@{\qquad}ll}
\toprule
\rowcolor{tableheader}
Setting & Value & Setting & Value\\
\midrule
Rollouts / prompt & 8 & Global batch & 128\\
Rollout temperature & 1.0 & Clip lower / upper & 0.20 / 0.28\\
Top-$p$ & 0.99 & Weight decay & $10^{-2}$\\
Prompt length & 4096 & Grad. norm & 1.0\\
Response length & 8192 & Precision & BF16\\
Vision tower & Trainable & Sharding & Full FSDP\\
Rollout batch & 384 & Epochs & 2\\
Learning rate & $10^{-6}$ & Bisection steps & 32\\
\bottomrule
\end{tabular}
\end{table}

\subsection{Baseline implementation}
GRPO uses group-relative reward normalization with the same asymmetric clipping as the primary trainer. DAPO adopts its sampling and clipping refinements without adding an auxiliary value model. PAPO is instantiated with its perception-aware objective on original and corrupted image pairs \citep{wang2026papo}. VPPO uses token visual dependency to shape trajectory advantages and select pivotal token gradients \citep{huang2026vppo}; PGPO applies token-level visual-dependency advantage reshaping with its mass-preserving design \citep{ye2026pgpo}; PRPO uses bounded dynamic advantage reshaping \citep{li2026prpo}; VL-Calibration combines visual certainty and reasoning confidence \citep{xiao2026vlcalibration}; and VGPO applies intra- and inter-trajectory visual-guidance reweighting \citep{wang2026vgpo}. For objective-side comparisons, all methods share the same rollout samples, verifier, base learning rate, response limits, and optimizer so that the dominant difference is the credit mechanism.

\subsection{Evaluation protocol}
The seven in-domain benchmarks use their official answer-extraction procedures and exact-match scoring. MathVerse, DynaMath, MMK12, Geo3K, and VisNumBench emphasize visual mathematical, geometric, or numerical reasoning; LogicVista stresses visual logical inference; MMMU-Pro evaluates multidisciplinary multimodal understanding. Out-of-domain tests use A-OKVQA \citep{schwenk2022aokvqa}, SimpleVQA \citep{cheng2025simplevqa}, ChartQA \citep{masry2022chartqa}, InfographicVQA \citep{mathew2022infographicvqa}, and DocVQA \citep{mathew2021docvqa}. Corruption diagnostics are generated from the evaluation images only and do not alter the ground-truth answer or textual question.

\subsection{Credit diagnostics}
Support flicker compares matched top-$\rho$ active sets under paired mild visual interventions. If $S_i$ and $\widetilde S_i$ are the active sets induced by two interventions, flicker is $1-|S_i\cap\widetilde S_i|/|S_i\cup\widetilde S_i|$. Credit continuity is the cosine similarity between the dense allocation fields. Gradient CV is the coefficient of variation of per-minibatch gradient norms over the final 20\% of updates. Entropy variance is computed over the same window from the policy's generated-token entropy. These diagnostics are measured on held-out trajectories and do not enter the training objective.

\subsection{Full Qwen3-VL-8B Baseline Comparison}

To verify that the main improvement remains competitive against a broader set of multimodal RLVR objectives, we extend the baseline pool beyond the compact comparison in the main text. Qwen3-VL-8B is trained on ViRL39K and evaluated on the same seven benchmarks with Instruction, GRPO, DAPO, Vision-R1, MM-Eureka, PAPO, PGPO, PRPO, VPPO, and \method{} under the protocol in Appendix~\ref{app:setup}.

\begin{table}[h]
\caption{Extended comparison on Qwen3-VL-8B. Values are accuracy@8 (\%); Avg. denotes the macro-average over all seven benchmarks. Bold and underlined values denote the best and second-best results, respectively.}
\label{tab:main_full}
\centering
\scriptsize
\setlength{\tabcolsep}{2.0pt}
\begin{tabular}{lcccccccc}
\toprule
\rowcolor{tableheader}
Method & MathV. & Dyna & MMK12 & Geo3K & VisNum & Logic & MMMU-P & Avg.\\
\midrule

Instruction
& 67.7
& 72.2
& 62.0
& 57.5
& 25.0
& 50.1
& 33.7
& 52.6\\

GRPO
& 76.5
& 75.0
& 71.2
& 64.4
& 28.7
& 56.0
& 42.8
& 59.2\\

DAPO
& 78.2
& 76.1
& 73.0
& 66.0
& 28.4
& 60.2
& 46.1
& 61.1\\

Vision-R1
& 79.0
& 76.8
& 74.1
& 66.8
& 28.9
& 58.0
& 50.7
& 62.0\\

MM-Eureka
& 80.1
& 78.5
& 76.0
& \underline{72.3}
& 29.5
& 61.1
& 52.1
& 64.2\\

PAPO
& 82.1
& 79.0
& 78.7
& 70.8
& 29.4
& 61.4
& 51.0
& 64.6\\

PGPO
& 82.7
& 79.4
& 79.5
& 71.1
& \underline{30.0}
& 61.8
& 51.9
& 65.2\\

PRPO
& 82.5
& 79.2
& 79.0
& 71.0
& 29.6
& 61.6
& 51.5
& 64.9\\

VPPO
& \underline{83.8}
& \underline{80.0}
& \underline{80.1}
& 71.9
& 29.1
& \underline{62.3}
& \underline{52.4}
& \underline{65.7}\\

\rowcolor{rvcorow}
\method{}
& \textbf{84.4}
& \textbf{80.4}
& \textbf{80.6}
& \textbf{72.5}
& \textbf{30.5}
& \textbf{64.1}
& \textbf{55.4}
& \textbf{66.8}\\

\bottomrule
\end{tabular}
\end{table}











Table~\ref{tab:main_full} shows that RVCO achieves the strongest overall
performance across the extended baseline pool. The improvement remains
consistent across mathematical reasoning, logical reasoning, and
multidisciplinary evaluation. The method obtains the best result on all
seven benchmarks and the highest overall average.

\section{Extended Credit Diagnostics}
\label{app:dynamics}
To separate support preservation from simple smoothing or changes in total credit mass, we first measure the geometry of the learned credit fields directly. On held-out Qwen3-VL-8B trajectories, Hard Top-$\rho$, independent sigmoid weighting, the entropic gate with $\beta=0$, and \method{} are evaluated under paired mild interventions using continuity, support overlap, weight Gini, and mass error.

\begin{table}[h]
\caption{Evidence-retention diagnostics on held-out trajectories. Credit continuity is cosine similarity between allocation fields under paired mild interventions; support overlap uses matched top-$\rho$ sets.}
\label{tab:diagnostics}
\centering
\small
\begin{tabular}{lcccc}
\toprule
\rowcolor{tableheader}
Method & Continuity $\uparrow$ & Support overlap $\uparrow$ & Weight Gini $\downarrow$ & Mass error $\downarrow$\\
\midrule
Hard Top-$\rho$ & 0.184 & 0.621 & 0.604 & 0.000\\
Independent sigmoid & 0.251 & 0.652 & 0.421 & 0.087\\
Entropic gate ($\beta=0$) & 0.267 & 0.638 & 0.468 & 0.000\\
\rowcolor{rvcorow} \method{} & \textbf{0.281} & \textbf{0.807} & \textbf{0.392} & \textbf{0.000}\\
\bottomrule
\end{tabular}
\end{table}

Table~\ref{tab:diagnostics} shows that independent sigmoid weighting improves continuity to 0.251 but incurs 0.087 mass error, whereas Hard Top-$\rho$ preserves mass exactly but has only 0.184 continuity. \method{} combines exact mass with the highest continuity (0.281), the highest support overlap (0.807), and the lowest weight Gini (0.392). The residual correction therefore improves support without changing the total routed mass.

To examine how this support difference evolves during optimization, we next track matched support overlap throughout training. The analysis uses Qwen3-VL-8B held-out trajectories and compares Hard Top-$\rho$, the entropic gate with $\beta=0$, and \method{} at matched training checkpoints under the same paired perturbations.

\begin{figure}[h]
\centering
\includegraphics[width=.82\linewidth]{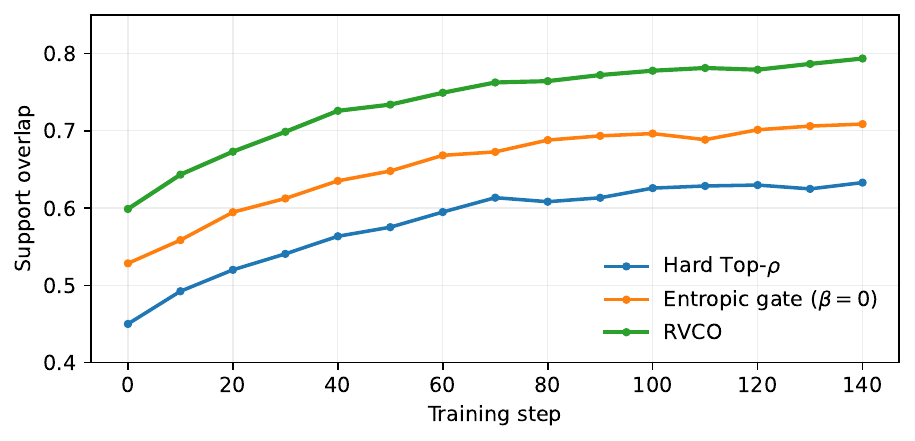}
\caption{Support overlap during training under matched perturbations. The residual route maintains a larger stable intersection of credited positions as evidence becomes sharper late in training.}
\label{fig:creditgeometry}
\end{figure}

Figure~\ref{fig:creditgeometry} shows that the three methods are closer early in training, but the gap widens as optimization proceeds. \method{} retains the highest overlap in the late stage, while Hard Top-$\rho$ remains substantially lower and the entropic-only variant lies between them. This trajectory-level pattern matches the final-window stability results in Table~\ref{tab:stability}.

To determine whether improved support stability is accompanied by better optimization progress, we also track held-out reward over training. Qwen3-VL-8B is trained with PGPO, VPPO, and \method{} under the shared protocol, and the dashed marker denotes the same late-stage window used for Table~\ref{tab:stability}.

\begin{figure}[h]
\centering
\includegraphics[width=.82\linewidth]{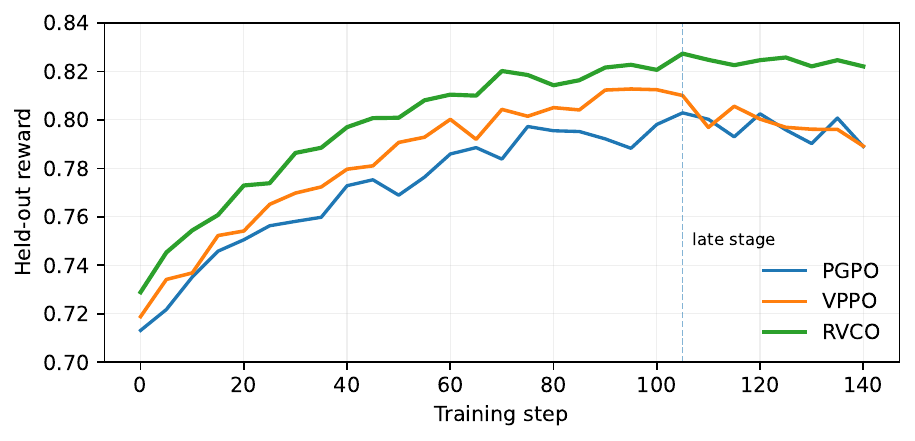}
\caption{Held-out reward across training for representative perception-aware methods. The vertical marker denotes the late-stage window used for stability diagnostics.}
\label{fig:stabilitycurve}
\end{figure}

Figure~\ref{fig:stabilitycurve} shows similar early gains for all three methods, followed by clearer separation after the late-stage marker. \method{} continues to increase toward the highest held-out reward, while VPPO bends downward and PGPO saturates earlier. This behavior is consistent with the lower entropy variance and gradient CV reported for the complete method.

To test the role of routing sharpness directly, we sweep the gate temperature over a 16$\times$ range. Qwen3-VL-4B is trained with $\tau\in\{0.03,0.06,0.12,0.24,0.48\}$ while all other settings remain fixed, and the training score is recorded through 140 steps.

\begin{figure}[h]
\centering
\includegraphics[width=.86\linewidth]{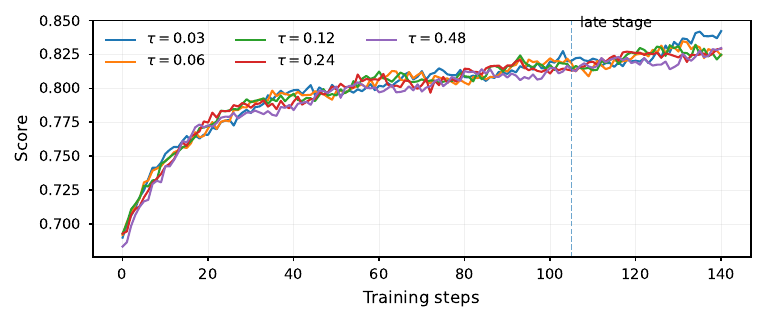}
\caption{Training dynamics over a 16$\times$ range of routing temperatures on Qwen3-VL-4B. Moderate temperatures preserve evidence contrast without abrupt late-stage changes.}
\label{fig:tempdyn}
\end{figure}

Figure~\ref{fig:tempdyn} shows that $\tau=0.06$, $0.12$, and $0.24$ follow closely matched trajectories, while the sharpest setting $\tau=0.03$ becomes more variable in the late stage and $\tau=0.48$ is slightly more conservative. The result supports a broad intermediate regime rather than a single narrow optimum.

To provide the corresponding validation view, we repeat the one-at-a-time sweep of $\beta$, $\rho$, and $\tau$ used in the main text. The experiment uses Qwen3-VL-8B on ViRL39K and reports training and MathVerse accuracy with the other parameters fixed at their defaults.

\begin{figure}[h]
\centering
\includegraphics[width=.88\linewidth]{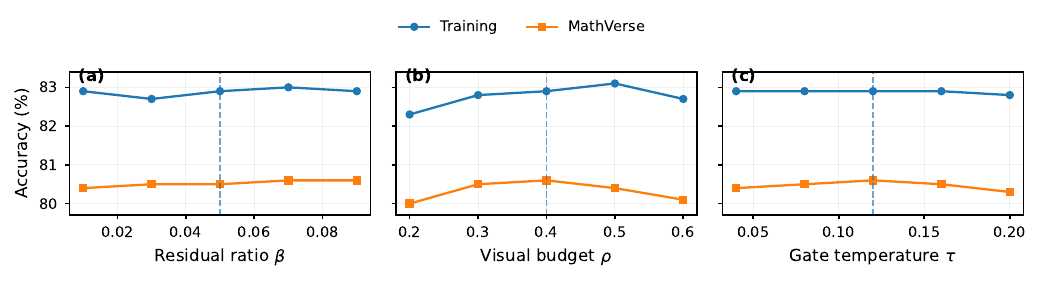}
\caption{Sensitivity to residual ratio $\beta$, visual budget $\rho$,and gate temperature $\tau$ on Qwen3-VL-8B. The dashed line marks the default.}
\label{fig:hyperappendix}
\end{figure}

Figure~\ref{fig:hyperappendix} confirms the same broad plateau reported in the main text: performance changes little across the $\beta$ and $\tau$ ranges, while $\rho=0.4$ gives the strongest joint validation point. This agreement across training and validation metrics indicates that the default configuration is not selected from an isolated spike.

\section{Out-of-Domain Generalization}
\paragraph{Out-of-Domain Generalization.}
To test whether reasoning-oriented credit optimization preserves visual-language capability outside the training-style reasoning suite, we evaluate unseen VQA and document tasks. Qwen3-VL-4B and Qwen3-VL-8B are evaluated on A-OKVQA, SimpleVQA, ChartQA, InfoVQA, and DocVQA using each benchmark's official primary metric, with the instruction model, GRPO, and \method{} compared under the same decoding protocol.

\begin{table}[H]
\caption{Out-of-domain visual question answering and document understanding. Values use each benchmark's official primary metric (\%).}
\label{tab:ood}
\centering
\scriptsize
\setlength{\tabcolsep}{3.5pt}
\begin{tabular}{llcccccc}
\toprule
\rowcolor{tableheader}
Backbone & Method & A-OKVQA & SimpleVQA & ChartQA & InfoVQA & DocVQA & Avg.\\
\midrule
\multirow{3}{*}{Qwen3-VL-8B} & Instruction & 87.5 & 43.9 & 82.4 & 55.6 & 89.7 & 71.8\\
& GRPO & 87.3 & 42.7 & 82.8 & 55.1 & 89.1 & 71.4\\
\rowcolor{rvcorow} & \method{} & \textbf{88.7} & \textbf{47.8} & \textbf{84.9} & \textbf{59.4} & \textbf{91.0} & \textbf{74.4}\\
\midrule
\multirow{3}{*}{Qwen3-VL-4B} & Instruction & 86.3 & 38.4 & 79.6 & 51.2 & 87.1 & 68.5\\
& GRPO & 86.2 & 37.1 & 79.9 & 50.8 & 86.5 & 68.1\\
\rowcolor{rvcorow} & \method{} & \textbf{87.3} & \textbf{42.0} & \textbf{82.1} & \textbf{55.0} & \textbf{88.4} & \textbf{71.0}\\
\bottomrule
\end{tabular}
\end{table}

Table~\ref{tab:ood} shows that \method{} raises the five-task average from 71.8 to 74.4 on Qwen3-VL-8B and from 68.5 to 71.0 on Qwen3-VL-4B relative to the instruction models. On the 8B backbone, the largest gains are on SimpleVQA (+3.9) and InfoVQA (+3.8), while DocVQA also improves by 1.3 points. The improvement therefore extends beyond the in-domain reasoning benchmarks rather than being confined to their answer format.

\section{Response-Length Analysis}
To test whether preserving weak but nonzero credit becomes more important as reasoning trajectories grow longer, we stratify evaluation by generated-response length. Qwen3-VL-8B is evaluated on the seven-task suite with GRPO, VPPO, and \method{}, and responses are grouped into $\le512$, 513--2048, and $>2048$ token buckets after answer extraction.

\begin{table}[h]
\caption{Accuracy by generated-response length on the seven-task reasoning suite. Length buckets are measured after answer extraction.}
\label{tab:length}
\centering
\small
\begin{tabular}{lcccc}
\toprule
\rowcolor{tableheader}
Length bucket & GRPO & VPPO & \method{} & Gain over VPPO\\
\midrule
$\le512$ tokens & 64.9 & 69.1 & \textbf{70.4} & +1.3\\
513--2048 tokens & 59.5 & 66.0 & \textbf{67.6} & +1.6\\
$>2048$ tokens & 53.2 & 60.6 & \textbf{63.0} & +2.4\\
\bottomrule
\end{tabular}
\end{table}

Table~\ref{tab:length} shows that \method{} improves VPPO by 1.3 points for responses up to 512 tokens, 1.6 points for 513--2048 tokens, and 2.4 points beyond 2048 tokens. Although accuracy falls with length for every method, the increasing margin indicates that full-support routing is most useful when the reasoning path contains more intermediate and revisited states.

\section{Rollout Group and Verifier Sensitivity}
To determine whether the token router depends on a particular group-relative advantage resolution, we vary rollout-group size. Qwen3-VL-8B is trained with 4, 8, and 16 responses per prompt while approximately matching the total generated-token budget through gradient accumulation; GRPO, VPPO, and \method{} use the same settings at each group size.

\begin{figure}[h]
\centering
\includegraphics[width=.8\linewidth]{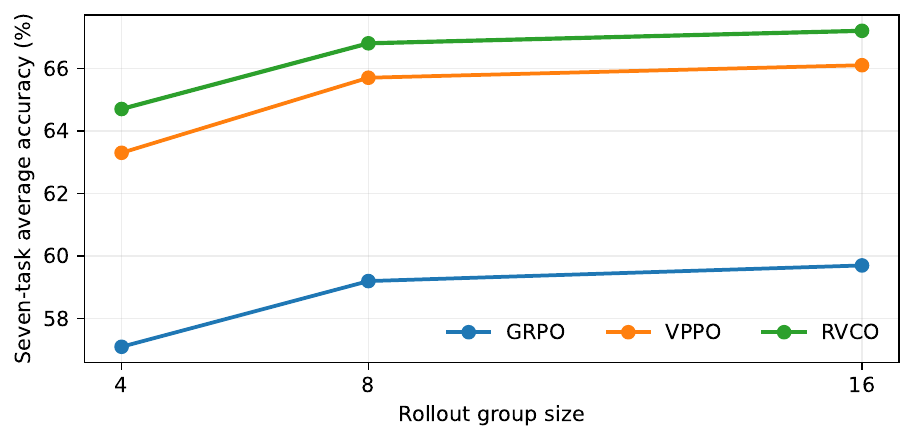}
\caption{Sensitivity to rollout-group size under approximately matched generated-token budgets.}
\label{fig:groupsize}
\end{figure}

Figure~\ref{fig:groupsize} shows that \method{} improves from 64.7 to 67.2 as the group size increases and remains 1.1--1.4 points above VPPO at every setting. The stable margin indicates that token-level conserved routing complements, rather than replaces, the variance reduction obtained from larger rollout groups.

To test sensitivity to imperfect outcome supervision, we next corrupt the verifier labels directly. On Qwen3-VL-8B, binary rewards are flipped at 0\%, 2\%, 5\%, and 10\% before group-relative normalization; GRPO, VPPO, and \method{} receive the same corrupted reward stream at each flip rate.

\begin{figure}[h]
\centering
\includegraphics[width=.8\linewidth]{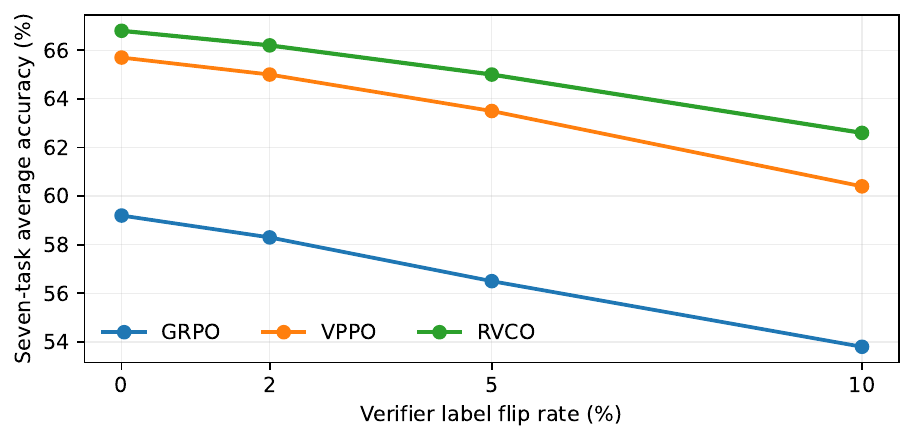}
\caption{Sensitivity to controlled verifier label noise. All methods use the same corrupted reward labels at each flip rate.}
\label{fig:verifiernoise}
\end{figure}

Figure~\ref{fig:verifiernoise} shows the expected decline for all methods as label noise grows, but \method{} remains above VPPO by 1.1 points at 0\% flips and 2.2 points at 10\% flips. The method does not correct erroneous rewards; the persistent margin instead shows that the credit operator remains beneficial when the sequence-level signal is moderately noisy.

\section{Intervention Design and Corruption Severity}
To determine whether the evidence coordinate depends on one particular visual perturbation, we compare several intervention operators. Qwen3-VL-8B is trained on ViRL39K with random patch masking, Gaussian noise, Gaussian blur, random cutout, or grayscale attenuation, and each variant is evaluated by seven-task average accuracy, credit continuity, and wall-clock overhead relative to GRPO.

\begin{table}[h]
\caption{Sensitivity to the visual intervention used to construct evidence coordinates on Qwen3-VL-8B. Overhead is relative to GRPO wall-clock time.}
\label{tab:intervention}
\centering
\small
\begin{tabular}{lccc}
\toprule
\rowcolor{tableheader}
Intervention & Avg. acc. $\uparrow$ & Continuity $\uparrow$ & Overhead\\
\midrule
Random patch mask (default) & \textbf{66.8} & \textbf{0.281} & 4.3\%\\
Gaussian noise & 66.3 & 0.275 & 4.1\%\\
Gaussian blur & 66.0 & 0.270 & 4.0\%\\
Random cutout & 66.1 & 0.272 & 4.2\%\\
Grayscale attenuation & 65.4 & 0.264 & 3.9\%\\
\bottomrule
\end{tabular}
\end{table}

Table~\ref{tab:intervention} shows that random patch masking gives the best average accuracy (66.8) and continuity (0.281), but Gaussian noise, blur, and cutout remain within 0.5--0.8 accuracy points of the default. All five interventions add about 4\% wall-clock overhead. The method therefore does not depend on a single corruption family, although localized patch masking provides the strongest overall result.

To test robustness beyond a single corruption operating point, we next vary the severity of patch corruption continuously. Qwen3-VL-8B is evaluated at 0\%, 10\%, 20\%, 30\%, and 40\% patch corruption with GRPO, VPPO, and \method{} under identical decoding and answer extraction.

\begin{figure}[h]
\centering
\includegraphics[width=.82\linewidth]{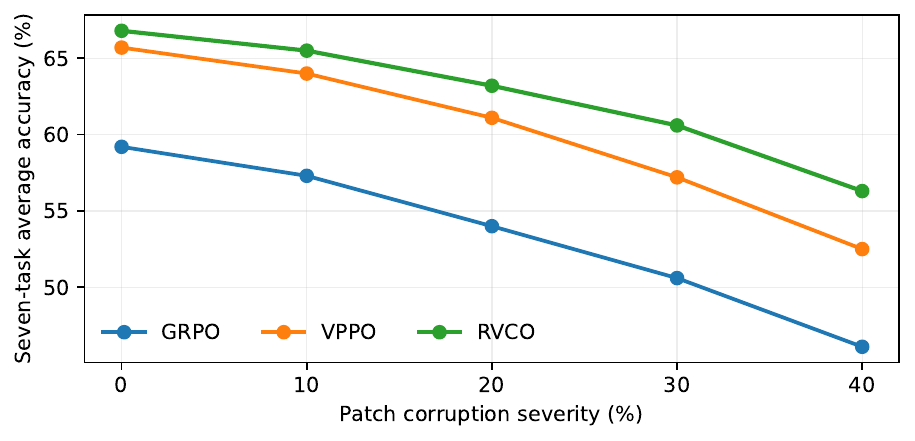}
\caption{Accuracy under increasing patch corruption severity. The full-support field degrades more gradually as the visual signal becomes incomplete.}
\label{fig:robustseverity}
\end{figure}

Figure~\ref{fig:robustseverity} shows that \method{} declines from 66.8 to 56.3 as corruption increases to 40\%, while VPPO falls from 65.7 to 52.5. The margin therefore grows from 1.1 points without corruption to 3.8 points at the highest severity. Full-support routing does not recover missing pixels, but it degrades more gradually when the visual evidence becomes diffuse.

\section{Data Efficiency and Allocation Case Study}
To verify the low-data trend independently of the main table, we plot reasoning accuracy across the full ViRL39K fraction sweep. Qwen3-VL-8B is trained with 25\%, 50\%, 75\%, and 100\% of the corpus using GRPO, VPPO, and \method{} under the same optimizer and seven-task evaluation protocol.

\begin{figure}[h]
\centering
\includegraphics[width=.82\linewidth]{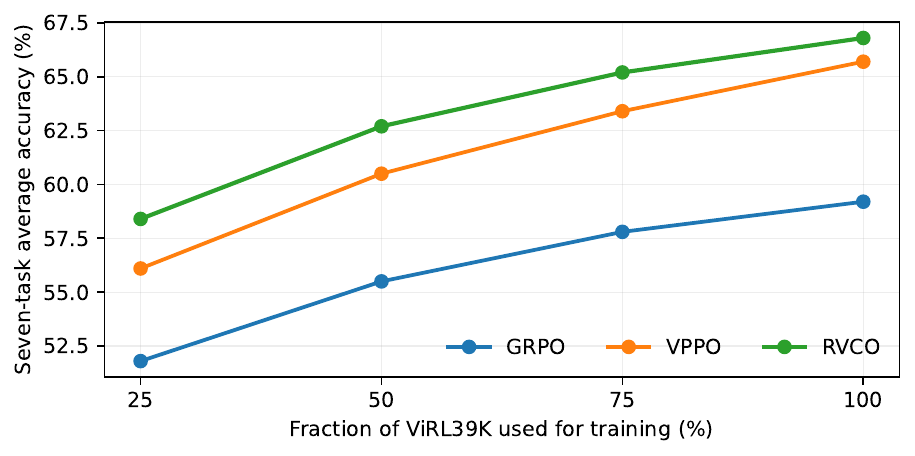}
\caption{Reasoning accuracy as the ViRL39K training fraction increases. The proposed field retains an advantage throughout the data range.}
\label{fig:dataeff}
\end{figure}

Figure~\ref{fig:dataeff} shows that \method{} rises from 58.4 at 25\% data to 66.8 at full data and remains above VPPO throughout. The gain over VPPO decreases from 2.3 points at 25\% data to 1.1 points at full data, while accuracy remains monotonic for all methods. The larger low-data margin indicates that conserved routing is especially useful when verifiable supervision is limited.

To inspect how the same fixed budget is distributed at token level, we also perform a paired qualitative allocation analysis. A Qwen3-VL-8B response is held fixed and scored under Hard Top-$\rho$ and \method{} with the same 40\% total credit budget, so the comparison changes only the allocation rule rather than the generated text or reward.

\begin{figure}[h]
\centering
\includegraphics[width=.9\linewidth]{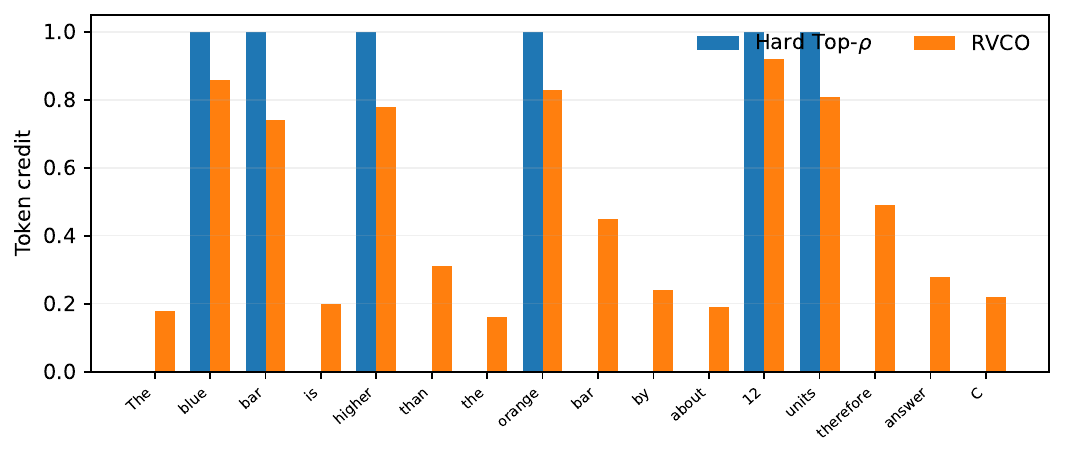}
\caption{Representative token-credit allocation for a visual comparison response under the same 40\% total budget. Hard routing alternates between unit and zero credit, whereas \method{} preserves graded support around visually grounded and connective tokens.}
\label{fig:tokencase}
\end{figure}

Figure~\ref{fig:tokencase} shows that the hard selector alternates between unit and zero weights, including abrupt changes around adjacent response tokens. \method{} assigns the largest weights to visually grounded content while retaining smaller positive weights on connective tokens under the same total budget. The example therefore illustrates full support without uniform credit: evidence-bearing tokens remain prioritized, but neighboring reasoning positions are not deleted from the update.

\section{Additional Implementation Notes}
The credit computation is implemented after original and intervened token log-probabilities have been materialized. Valid-token masks are reused from the PPO batch; quantile statistics ignore padding and prompt tokens. The bisection interval is initialized from the minimum and maximum calibrated evidence coordinates with a fixed numerical margin, which avoids a per-response search for custom bounds. Thirty-two iterations are sufficient for the mass error to fall below the precision of the subsequent BF16 policy update.

For distributed execution, the intervened forward pass can be scheduled alongside reward-side diagnostics because it does not require policy gradients. The resulting $d_{i,t}$ and $z_{i,t}$ tensors are detached before the PPO minibatch is formed. This keeps the computational graph identical to the base policy update and avoids an additional backward pass through the intervention branch. The extra persistent storage is one scalar evidence coordinate and one scalar weight per generated token; temporary full-vocabulary logits can be released immediately after the sampled-token log-probability is gathered.

The budget ratio $\rho$ is applied to valid response positions rather than padded sequence length. For very short responses, $m_i=\lceil\rho T_i\rceil$ ensures at least one unit of routed mass. For long responses, the linear scaling keeps the average gradient contribution comparable across examples. Residual support does not change this amount because the analytic correction is computed from $T_i$, $m_i$, and $\beta$ in closed form.

\section{Limitations and Broader Impact}
\textbf{Limitations.}
\method{} requires an additional likelihood evaluation under a visually
intervened input, introducing modest overhead relative to outcome-only
GRPO. The evidence response measures conditional sensitivity to a chosen
intervention distribution rather than a formal causal effect, and the
intervention may underrepresent visual evidence that is robust to masking
or distributed across large image regions. The visual budget $\rho$, gate
temperature $\tau$, and residual ratio $\beta$ are fixed throughout
training; adaptive schedules or response-dependent settings may be
beneficial for unusually long trajectories or tasks in which visual
dependence is highly sparse. The present study focuses on static-image
tasks with verifiable outcomes; extending conserved evidence routing to
video, embodied interaction, multi-image settings, and open-ended
preference-based rewards remains future work.


\end{document}

%% file: math_commands.tex
\usepackage{amsmath,amsfonts,bm}

\def\eqref#1{equation~\ref{#1}}

\def\1{\bm{1}}

\DeclareMathAlphabet{\mathsfit}{\encodingdefault}{\sfdefault}{m}{sl}
\SetMathAlphabet{\mathsfit}{bold}{\encodingdefault}{\sfdefault}{bx}{n}

\DeclareMathOperator*{\argmax}{arg\,max}

%% file: theory.tex
\section{Optimization Properties}
\label{app:proof}

The constrained entropic routing step used in \method{} is a
temperature-scaled instance of the Limited Multi-Label (LML) projection
\citep{amos2019limited}. It also belongs to the broader class of
regularized prediction maps \citep{blondel2020learning}. Related
differentiable top-$k$ constructions can be obtained through entropic
optimal transport \citep{xie2020differentiable}. We therefore do not
attribute novelty to the projection primitive itself. We restate its
standard optimization properties for completeness and then analyze the
residual-support and exact-conservation operations used to construct the
credit field in \method{}.
For a response of length $T$ and budget $m$, define the capped
fixed-mass polytope
\begin{equation}
\mathcal{L}_{T,m}
=
\left\{
g\in[0,1]^T:
\mathbf{1}^{\top}g=m
\right\},
\label{eq:lml_polytope}
\end{equation}
and the binary entropy
\begin{equation}
H_{\mathrm b}(g)
=
\sum_{t=1}^{T}h(g_t),
\qquad
h(a)
=
-a\log a-(1-a)\log(1-a).
\label{eq:binary_entropy_app}
\end{equation}
The routing map is
\begin{equation}
g_\tau(z)
=
\argmax_{g\in\mathcal{L}_{T,m}}
\left\{
z^\top g+\tau H_{\mathrm b}(g)
\right\},
\qquad
\tau>0.
\label{eq:entropic_router_app}
\end{equation}
Dividing the objective by $\tau$ gives
\begin{equation}
g_\tau(z)
=
\argmax_{g\in\mathcal{L}_{T,m}}
\left\{
\left(\frac{z}{\tau}\right)^\top g
+
H_{\mathrm b}(g)
\right\},
\label{eq:lml_equivalence}
\end{equation}
which is exactly the LML projection of the temperature-scaled score
vector $z/\tau$ under the binary-entropy regularizer
\citep{amos2019limited}. Equivalently, with
\begin{equation}
\Omega_{\tau,m}(g)
=
-\tau H_{\mathrm b}(g)
+
I_{\mathcal{L}_{T,m}}(g),
\end{equation}
where $I_{\mathcal{L}_{T,m}}$ is the convex indicator of
$\mathcal{L}_{T,m}$, Equation~\eqref{eq:entropic_router_app} has the
regularized-prediction form
\begin{equation}
g_\tau(z)
=
\argmax_g
\left\{
z^\top g-\Omega_{\tau,m}(g)
\right\}.
\end{equation}

\subsection{Existence, uniqueness, and scalar-dual solution}

Consider first the non-degenerate case $0<m<T$.
The feasible set $\mathcal{L}_{T,m}$ is compact and convex, and the
objective in Equation~\eqref{eq:entropic_router_app} is continuous.
Hence a maximizer exists. Binary entropy is strictly concave, so the
objective is strictly concave on the feasible affine set. The maximizer
is therefore unique.

Moreover,
\begin{equation}
h'(a)
=
\log\frac{1-a}{a}
\end{equation}
diverges to $+\infty$ as $a\rightarrow0^{+}$ and to $-\infty$ as
$a\rightarrow1^{-}$. Thus the optimum lies in the relative interior for
$0<m<T$, and the box constraints are inactive at finite $\tau$.
Introducing a multiplier $\lambda$ for
$\mathbf{1}^{\top}g=m$, the Lagrangian is
\begin{equation}
\mathcal{J}(g,\lambda)
=
z^\top g
+
\tau H_{\mathrm b}(g)
-
\lambda(\mathbf{1}^{\top}g-m).
\end{equation}
Stationarity gives
\begin{equation}
z_t
+
\tau\log\frac{1-g_t}{g_t}
-
\lambda
=
0,
\end{equation}
and hence
\begin{equation}
g_t
=
\sigma\!\left(
\frac{z_t-\lambda}{\tau}
\right),
\qquad
\sigma(x)=\frac{1}{1+\exp(-x)}.
\label{eq:lml_sigmoid}
\end{equation}
The equality constraint reduces the complete optimization problem to
the scalar equation
\begin{equation}
F(\lambda)
=
\sum_{t=1}^{T}
\sigma\!\left(
\frac{z_t-\lambda}{\tau}
\right)
=
m.
\label{eq:dual_mass}
\end{equation}
Its derivative is
\begin{equation}
F'(\lambda)
=
-\frac{1}{\tau}
\sum_{t=1}^{T}
g_t(1-g_t)
<
0.
\label{eq:F_derivative}
\end{equation}
Furthermore,
\begin{equation}
\lim_{\lambda\rightarrow-\infty}F(\lambda)=T,
\qquad
\lim_{\lambda\rightarrow+\infty}F(\lambda)=0.
\end{equation}
Therefore Equation~\eqref{eq:dual_mass} has exactly one solution.

The same result follows directly from the one-dimensional dual. Taking
the supremum of the Lagrangian over each coordinate gives
\begin{equation}
\phi(\lambda)
=
m\lambda
+
\tau
\sum_{t=1}^{T}
\log
\left(
1+
\exp\frac{z_t-\lambda}{\tau}
\right).
\label{eq:scalar_dual}
\end{equation}
Its first and second derivatives are
\begin{align}
\phi'(\lambda)
&=
m-F(\lambda),\\
\phi''(\lambda)
&=
\frac{1}{\tau}
\sum_{t=1}^{T}
g_t(1-g_t)
>
0.
\end{align}
Thus $\phi$ is strictly convex and has a unique minimizer. Bisection on
Equation~\eqref{eq:dual_mass} therefore recovers the unique primal
solution. This is the scalar-dual structure of the LML projection
\citep{amos2019limited}.

For the boundary case $m=T$, the feasible set contains only
$g=\mathbf{1}$, so uniqueness and conservation hold trivially.

\subsection{Shift invariance and order preservation}

The common dual variable makes the invariances of the routing map
explicit. Consider a shifted score vector
\begin{equation}
z'
=
z+a\mathbf{1},
\end{equation}
for an arbitrary scalar $a$. If $\lambda$ satisfies
Equation~\eqref{eq:dual_mass} for $z$, then
\begin{equation}
\lambda'
=
\lambda+a
\end{equation}
satisfies the same mass equation for $z'$, because
\begin{equation}
\frac{z'_t-\lambda'}{\tau}
=
\frac{z_t-\lambda}{\tau}.
\end{equation}
Consequently,
\begin{equation}
g_\tau(z+a\mathbf{1})
=
g_\tau(z).
\label{eq:shift_invariance}
\end{equation}
The router therefore depends only on relative evidence scores and not
on an arbitrary response-level offset.

Order preservation follows immediately from the strict monotonicity of
the sigmoid. For any two coordinates $r$ and $s$,
\begin{equation}
z_r>z_s
\quad\Longleftrightarrow\quad
\frac{z_r-\lambda}{\tau}
>
\frac{z_s-\lambda}{\tau}
\quad\Longleftrightarrow\quad
g_r>g_s.
\label{eq:order_preservation}
\end{equation}
Thus entropic smoothing changes the magnitude of the allocation but not
the ranking induced by the evidence scores.

\subsection{Jacobian and finite-temperature stability}

The scalar-dual form also yields an explicit Jacobian. Define
\begin{equation}
q_t
=
g_t(1-g_t),
\qquad
Q
=
\sum_{r=1}^{T}q_r.
\end{equation}
Differentiating the mass constraint
$\sum_t g_t=m$ with respect to $z_s$ gives
\begin{equation}
\frac{\partial\lambda}{\partial z_s}
=
\frac{q_s}{Q}.
\label{eq:lambda_derivative}
\end{equation}
Using Equation~\eqref{eq:lml_sigmoid},
\begin{equation}
\frac{\partial g_t}{\partial z_s}
=
\frac{q_t}{\tau}
\left(
\mathbb{I}[t=s]
-
\frac{q_s}{Q}
\right).
\label{eq:gate_jacobian}
\end{equation}
In matrix form,
\begin{equation}
J_g
=
\frac{1}{\tau}
\left(
D_q-\frac{qq^\top}{Q}
\right),
\qquad
D_q=\operatorname{diag}(q).
\label{eq:gate_jacobian_matrix}
\end{equation}
For any $v\in\mathbb{R}^{T}$,
\begin{align}
v^\top J_g v
=
\frac{1}{\tau}
\left[
\sum_{t=1}^{T}q_t v_t^2
-
\frac{
\left(\sum_{t=1}^{T}q_t v_t\right)^2
}{Q}
\right]
\ge 0,
\label{eq:jacobian_psd}
\end{align}
where the inequality follows from Cauchy--Schwarz. Hence $J_g$ is
positive semidefinite. It also satisfies
\begin{equation}
J_g\mathbf{1}=0,
\end{equation}
which is the differential form of shift invariance and exact budget
conservation.

Because $0<q_t\le1/4$,
\begin{equation}
0
\preceq
J_g
\preceq
\frac{1}{\tau}D_q
\preceq
\frac{1}{4\tau}I,
\end{equation}
and therefore
\begin{equation}
\|J_g\|_2
\le
\frac{1}{4\tau}.
\label{eq:lipschitz_gate}
\end{equation}
Finite temperature thus produces a smooth allocation map with a
controlled local sensitivity. Decreasing $\tau$ sharpens the response
and increases sensitivity near rank boundaries.

Equation~\eqref{eq:gate_jacobian} also exposes the competition induced
by the fixed budget:
\begin{equation}
\frac{\partial g_t}{\partial z_t}
=
\frac{q_t}{\tau}
\left(1-\frac{q_t}{Q}\right)
>0,
\end{equation}
while for $s\neq t$,
\begin{equation}
\frac{\partial g_t}{\partial z_s}
=
-
\frac{q_tq_s}{\tau Q}
<0.
\end{equation}
Increasing one score therefore increases its own allocation while
slightly reducing the allocation available to the remaining positions.
This interaction is absent from independent sigmoid gating.

\subsection{Conserved residual support}

The method-specific credit field is obtained after the entropic
projection. For $0<\beta<1$, define
\begin{equation}
\bar g_t
=
\beta+(1-\beta)g_t.
\label{eq:residual_gate_app}
\end{equation}
Since $\sum_t g_t=m$,
\begin{equation}
\sum_{t=1}^{T}\bar g_t
=
\beta T+(1-\beta)m.
\end{equation}
The exact mass-correction factor is therefore
\begin{equation}
c
=
\frac{m}
{\beta T+(1-\beta)m},
\label{eq:mass_correction}
\end{equation}
and Equation~\ref{eq:residual} can be written as
\begin{equation}
w_t
=
c\,
\bigl[
\beta+(1-\beta)g_t
\bigr].
\label{eq:residual_expanded}
\end{equation}

The total routed credit is exactly conserved:
\begin{align}
\sum_{t=1}^{T}w_t
&=
c
\sum_{t=1}^{T}
\bigl[
\beta+(1-\beta)g_t
\bigr]\\
&=
\frac{m}
{\beta T+(1-\beta)m}
\left[
\beta T+(1-\beta)m
\right]\\
&=
m.
\label{eq:proof_conservation}
\end{align}
Moreover, $m\le T$ implies
\begin{equation}
0<c
=
\frac{m}
{\beta T+(1-\beta)m}
\le1.
\end{equation}
Since $0\le g_t\le1$,
\begin{equation}
c\beta
\le
w_t
\le
c
\le
1.
\label{eq:full_support_bounds}
\end{equation}
Hence every token receives strictly positive credit for $\beta>0$,
while no coefficient exceeds one. This establishes both full support
and boundedness.

The residual transformation is affine and strictly increasing in
$g_t$:
\begin{equation}
\frac{\partial w_t}{\partial g_t}
=
c(1-\beta)
>
0.
\end{equation}
It therefore preserves the ordering produced by the entropic router.
Combining this fact with Equations~\eqref{eq:shift_invariance} and
\eqref{eq:order_preservation} gives
\begin{equation}
z_r>z_s
\quad\Longrightarrow\quad
w_r>w_s,
\qquad
w(z+a\mathbf{1})=w(z).
\end{equation}

Because $c$ depends only on $(T,m,\beta)$, the Jacobian of the final
credit field is
\begin{equation}
J_w
=
c(1-\beta)J_g.
\label{eq:residual_jacobian}
\end{equation}
Consequently,
\begin{equation}
\|J_w\|_2
\le
\frac{c(1-\beta)}{4\tau}
<
\frac{1}{4\tau}.
\label{eq:residual_lipschitz}
\end{equation}
Residual support therefore retains the ranking and mass constraint of
the projected field while attenuating its local sensitivity.

\subsection{Sparse and uniform limits}

The temperature parameter continuously connects a hard budgeted
selection with a uniform allocation.

First consider $\tau\rightarrow0$. Since binary entropy is bounded on
the unit interval,
\begin{equation}
0
\le
H_{\mathrm b}(g)
\le
T\log 2.
\end{equation}
Hence
\begin{equation}
\sup_{g\in\mathcal{L}_{T,m}}
\left|
\tau H_{\mathrm b}(g)
\right|
\le
\tau T\log 2
\longrightarrow
0.
\end{equation}
The regularized objective therefore converges uniformly to the linear
program
\begin{equation}
\max_{g\in\mathcal{L}_{T,m}}
z^\top g.
\label{eq:hard_budget_lp}
\end{equation}

Let
\begin{equation}
z_{(1)}>z_{(2)}>\cdots>z_{(T)}
\end{equation}
denote distinct coordinates sorted in descending order. If $m$ is an
integer, Equation~\eqref{eq:hard_budget_lp} has the unique solution
\begin{equation}
g_{(t)}^{0}
=
\begin{cases}
1, & t\le m,\\
0, & t>m.
\end{cases}
\label{eq:hard_topm}
\end{equation}
Thus
\begin{equation}
g_\tau(z)
\longrightarrow
g^{0}
\qquad
\text{as }\tau\rightarrow0,
\end{equation}
which is the hard top-$m$ selector.

For completeness, if a relaxed non-integer budget
$m=k+r$ is used with $k=\lfloor m\rfloor$ and $0<r<1$, the limiting
capped solution is
\begin{equation}
g_{(t)}^{0}
=
\begin{cases}
1, & t\le k,\\
r, & t=k+1,\\
0, & t>k+1.
\end{cases}
\end{equation}
The integer-budget case reduces exactly to
Equation~\eqref{eq:hard_topm}.

Next consider the residual parameter. From
Equation~\eqref{eq:mass_correction},
\begin{equation}
\lim_{\beta\rightarrow0}c=1,
\end{equation}
and therefore
\begin{equation}
\lim_{\beta\rightarrow0}w_t
=
g_t.
\end{equation}
Combining the two limits gives the hard selector stated in
Proposition~1 for distinct scores and integer $m$:
\begin{equation}
\lim_{\substack{\tau\rightarrow0\\\beta\rightarrow0}}
w
=
g^{0}.
\label{eq:joint_sparse_limit}
\end{equation}

At the opposite extreme, $\tau\rightarrow\infty$ suppresses all
finite score differences. From
Equation~\eqref{eq:lml_sigmoid},
\begin{equation}
\operatorname{logit}(g_t)
-
\operatorname{logit}(g_s)
=
\frac{z_t-z_s}{\tau}
\longrightarrow
0.
\end{equation}
All coordinates therefore approach a common value. The mass constraint
determines this value uniquely:
\begin{equation}
g_t
\longrightarrow
\frac{m}{T}.
\end{equation}
Substituting this limit into
Equation~\eqref{eq:residual_expanded} yields
\begin{align}
w_t
&\longrightarrow
\frac{m}
{\beta T+(1-\beta)m}
\left[
\beta+(1-\beta)\frac{m}{T}
\right]\\
&=
\frac{m}{T}.
\end{align}
Thus both the projected gate and the final conserved field converge to
the same uniform allocation. The complete path from sparse selection to
uniform credit preserves total mass $m$.

\subsection{Exact conservation versus unit-mean normalization}

The distinction between exact mass conservation and unit-mean
normalization is algebraic rather than cosmetic. Let
\begin{equation}
a_t
=
\beta+(1-\beta)g_t,
\qquad
S
=
\sum_{t=1}^{T}a_t
=
\beta T+(1-\beta)m.
\end{equation}
Exact conservation uses
\begin{equation}
w_t^{\mathrm{mass}}
=
\frac{m}{S}a_t,
\qquad
\sum_{t=1}^{T}w_t^{\mathrm{mass}}
=
m.
\label{eq:mass_normalization}
\end{equation}
Unit-mean normalization instead gives
\begin{equation}
w_t^{\mathrm{mean}}
=
\frac{T}{S}a_t,
\qquad
\frac{1}{T}
\sum_{t=1}^{T}
w_t^{\mathrm{mean}}
=
1.
\label{eq:mean_normalization}
\end{equation}
The two fields have identical relative ordering, but their global mass
differs by the exact factor
\begin{equation}
w_t^{\mathrm{mean}}
=
\frac{T}{m}
w_t^{\mathrm{mass}}.
\label{eq:normalization_ratio}
\end{equation}
For $m=\rho T$, this becomes
\begin{equation}
w_t^{\mathrm{mean}}
=
\frac{1}{\rho}
w_t^{\mathrm{mass}}.
\end{equation}
Unit-mean normalization therefore fixes the total coefficient mass at
$T$, whereas the proposed correction fixes it at the prescribed routing
budget $m$. In a weighted policy objective, these choices induce
different global scales even when their token rankings are identical.

The role of exact conservation in \method{} is therefore specific: the
residual floor can restore support to low-ranked tokens without changing
the total credit assigned to the response. The allocation geometry is
inherited from the known entropic projection, while residual support and
analytic mass correction determine how that projection is converted
into a conserved credit field for multimodal RLVR.